\documentclass[11pt,a4paper]{article}
\usepackage[hyperref]{acl2018}
\usepackage{times}
\usepackage{latexsym}
\usepackage{url}
\usepackage{multirow}
\usepackage{mathtools}
\usepackage{floatrow}
\usepackage{subcaption}

\aclfinalcopy 

\title{Syntax Helps ELMo Understand Semantics: \\Is Syntax Still Relevant in a Deep Neural Architecture for SRL?}

\author{Emma Strubell \qquad Andrew McCallum \\
  College of Information and Computer Sciences \\
    University of Massachusetts Amherst \\
    {\tt \{strubell, mccallum\}@cs.umass.edu}}

\date{}

\begin{document}
\maketitle
\begin{abstract}
Do unsupervised methods for learning rich, contextualized token representations obviate the need for explicit modeling of linguistic structure in neural network models for semantic role labeling (SRL)? We address this question by incorporating the massively successful ELMo embeddings \citep{peters2018deep} into LISA \citep{strubell2018linguistically}, a strong, linguistically-informed neural network architecture for SRL. In experiments on the CoNLL-2005 shared task we find that though ELMo out-performs typical word embeddings, beginning to close the gap in F1 between LISA with predicted and gold syntactic parses, syntactically-informed models still out-perform syntax-free models when both use ELMo, especially on out-of-domain data. Our results suggest that linguistic structures are indeed still relevant in this golden age of deep learning for NLP.
\end{abstract}

\section{Introduction}
Many state-of-the-art NLP models are now ``end-to-end'' deep neural network architectures which eschew explicit linguistic structures as input in favor of operating directly on raw text \citep{ma2016end,lee2017end,tan2018deep}. Recently, \citet{peters2018deep} proposed a method for unsupervised learning of rich, contextually-encoded token representations which, when supplied as input word representations in end-to-end models, further increased these models' performance by up to 25\% across many NLP tasks. The immense success of these linguistically-agnostic models brings into question whether linguistic structures such as syntactic parse trees still provide any additional benefits in a deep neural network architecture for e.g.~semantic role labeling (SRL).

In this work, we aim to begin to answer this question by experimenting with incorporating the ELMo embeddings of \citet{peters2018deep} into LISA \citep{strubell2018linguistically}, a ``linguistically-informed'' deep neural network architecture for SRL which, when given weaker GloVe embeddings as inputs \citep{pennington2014glove}, has been shown to leverage syntax to out-perform a state-of-the-art, linguistically-agnostic end-to-end SRL model. 

In experiments on the CoNLL-2005 English SRL shared task, we find that, while the ELMo representations out-perform GloVe and begin to close the performance gap between LISA with predicted and gold syntactic parses, syntactically-informed models still out-perform syntax-free models, especially on out-of-domain data. Our results suggest that with the right modeling, incorporating linguistic structures can indeed further improve strong neural network models for NLP.

\section{Models}
We are interested in assessing whether linguistic information is still beneficial in addition to deep, contextualized ELMo word embeddings in a neural network model for SRL. Towards this end, our base model for experimentation is Linguistically-Informed Self-Attention (LISA) \citep{strubell2018linguistically}, a deep neural network model which uses multi-head self-attention in the style of \citet{vaswani2017attention} for multi-task learning \citep{caruana1993multitask} across SRL, predicate detection, part-of-speech tagging and syntactic parsing. Syntax is incorporated by training one self-attention head to attend to each token's syntactic head, allowing it to act as an oracle providing syntactic information to further layers used to predict semantic roles. We summarize the key points of LISA in \S\ref{sec:lisa}. 

\citet{strubell2018linguistically} showed that LISA out-performs syntax-free models when both use GloVe word embeddings as input, which, due to their availability, size and large training corpora, are typically used as input to end-to-end NLP models. In this work, we replace those token representations with ELMo representations to assess whether ELMo embeddings are sufficiently rich to obviate the need for explicit representations of syntax, or the model still benefits from syntactic information in addition to the rich ELMo encodings. In \S\ref{sec:glove} and \S\ref{sec:elmo} we summarize how GloVe and ELMo embeddings, respectively, are incorporated into this model.  

\subsection{LISA SRL model \label{sec:lisa}}

\subsubsection{Neural network token encoder}

The input to LISA is a sequence $\mathcal{X}$ of $T$ token representations $x_t$. The exact form of these representations when using GloVe embeddings is described in \S\ref{sec:glove}, and for ELMo described in \S\ref{sec:elmo}. Following \citet{vaswani2017attention}, we add a sinusoidal positional encoding to these vectors since the self-attention has no inherent mechanism for modeling token position.

These token representations are supplied to a series of $J$ multi-head self-attention layers similar to those that make up the encoder model of \citet{vaswani2017attention}. We denote the $j$th layer with the function $S^{(j)}(\cdot)$ and the output of that layer for token $t$ as $s_t^{(j)}$:
\begin{align}
s_t^{(j)} = S^{(j)}(s_t^{(j-1)})
\end{align}
Each $S^{(j)}(\cdot)$ consists of two components: (a) multi-head self-attention and (b) a convolutional layer. For brevity, we will detail (a) here as it is how we incorporate syntax into the model, but we leave the reader to refer to \citet{strubell2018linguistically} for more details on (b).

The multi-head self attention consists of $H$ attention heads, each of which learns a distinct attention function to attend to all of the tokens in the sequence. This self-attention is performed for each token for each head, and the results of the $H$ self-attentions are concatenated to form the final self-attended representation for each token. 

Specifically, consider the matrix $S^{(j-1)}$ of $T$ token representations at layer $j-1$. For each attention head $h$, we project this matrix into distinct key, value and query representations $K_h^{(j)}$, $V_h^{(j)}$ and $Q_h^{(j)}$ of dimensions $T\times d_k$, $T\times d_q$, and $T\times d_v$, respectively. We can then multiply $Q_h^{(j)}$ by $K_h^{(j)}$ to obtain a $T\times T$ matrix of attention weights $A_h^{(j)}$ between each pair of tokens in the sentence. Following \citet{vaswani2017attention} we perform scaled dot-product attention: We scale the weights by the inverse square root of their embedding dimension and normalize with the softmax function to produce a distinct distribution for each token over all the tokens in the sentence:
\begin{align}
A_h^{(j)} = \mathrm{softmax}(d_{k}^{-0.5}Q_h^{(j)}{K_h^{(j)}}^T)
\end{align}
These attention weights are then multiplied by $V_h^{(j)}$ for each token to obtain the self-attended token representations $M_h^{(j)}$:
\begin{align}
M_h^{(j)} = A_h^{(j)}V_h^{(j)}
\end{align}
Row $t$ of $M_h^{(j)}$, the self-attended representation for token $t$ at layer $j$, is thus the weighted sum with respect to $t$ (given by $A_h^{(j)}$) over the token representations in $V_h^{(j)}$. The representations for each attention head are concatenated, and this representation is fed through a convolutional layer to produce $s_t^{(j)}$. In all of our models, we use $J=4$, $H=8$ and $d_k=d_q=d_v=64$.


\subsubsection{Incorporating syntax}
LISA incorporates syntax by training one attention head to attend to each token's parent in a syntactic dependency parse tree. At layer $j_{p}$, $H-1$ heads are left to learn on their own to attend to relevant tokens in the sentence, while one head $h_{p}$ is trained with an auxiliary objective which encourages the head to put all attention weight on each token's syntactic parent. Denoting the entry of $A_{h_{p}}^{(j_{p})}$ corresponding to the attention from token $t$ to token $q$ as $a_{tq}$, then we model the probability that $q$ is the head of $t$ as: $P(q=\mathrm{head}(t) \mid \mathcal{X}) = a_{tq}$. Trained in this way, $A_{h_{p}}^{(j_{p})}$ emits a directed graph, where each token's syntactic parent is that which is assigned the highest attention weight. During training, this head's attention weights are set to match the gold parse: $A_{h_{p}}^{(j_{p})}$ is set to the adjacency matrix of the parse tree,\footnote{Roots are represented by self-loops.} allowing downstream layers to learn to use the parse information throughout training. In our experiments we set $j_{p}=3$.

In this way, LISA is trained to use $A_{h_{p}}^{(j_{p})}$ as an oracle providing parse information to downstream layers. This representation is flexible, allowing LISA to use its own predicted parse, or a parse produced by another dependency parser. Since LISA is trained to leverage gold parse information, as higher-accuracy dependency parses become available, they can be provided to LISA to improve SRL without requiring re-training of the SRL model.

\subsubsection{Predicting POS and predicates \label{sec:pos}}
LISA is also trained to predict parts-of-speech and predicates using hard parameter sharing \citep{caruana1993multitask}. At layer $j_{pos}$, the token representation $s_t^{(j_{pos})}$ is provided as features for a multi-class classifier into the joint label space of part-of-speech and (binary) predicate labels: For each part-of-speech tag which is the tag for a predicate in the training data, we add a tag of the form {\sc tag:predicate}. Locally-normalized probabilities are computed using the softmax function, and we minimize the sum of this loss term with the SRL and syntax losses. In our experiments we use $j_{pos}=2$.

\subsubsection{Predicting semantic roles}
LISA's final network representations $S^{(J)}$ are used to predict semantic roles. Each token's final representation $s_t^{(J)}$ is projected to distinct \emph{predicate} and \emph{role} representations $s_t^{pred}$ and $s_t^{role}$. Each predicted predicate\footnote{During training, semantic role predictions are conditioned on the gold predicates. At test time they are conditioned on LISA's predicted predicates (\S\ref{sec:pos}).} is scored against all other tokens' role representations to produce per-label scores for each predicate-token pair using a bilinear operator $U$. Per-label scores across semantic roles with respect to predicate $f$ and token $t$ are thus given by:
\begin{align}
s_{ft} = s_f^{pred} U s_t^{role}
\end{align}
With the locally-normalized probability of the correct role label $y_{ft}$ given by: $P(y_{ft} \mid \mathcal{X}) \propto \mathrm{softmax}(s_{ft})$. At test time, we use Viterbi decoding to enforce BIO constraints with fixed transition probabilities between tags obtained from the training data.

\subsection{GLoVe embedding model\label{sec:glove}}

The GloVe word embedding model \citep{pennington2014glove}, like word2vec's skip-gram and CBOW \citep{mikolov2013distributed} algorithms, is a shallow, log-bilinear embedding model for learning unsupervised representations of words based on the intuition that words which occur in similar contexts should have similar representations. GloVe Vectors are learned for each word in a fixed vocabulary by regressing on entries in the word co-occurrence matrix constructed from a large corpus: The dot product between two words' embeddings should equal the log probability of the words' co-occurrence in the data. We refer the reader to \citet{pennington2014glove} for a more detailed description of the model.

We incorporate pre-trained GloVe embeddings into our model following \citet{strubell2018linguistically}: We fix the pre-trained embeddings and add a learned word embedding representation to the pre-trained word vectors, following the intuition that fixing the pre-trained embeddings and learning a residual word representation keeps words observed during training from drifting too far away from the pre-trained representations of unobserved words. We then feed these representations through $K$ width-3 convolutional layers with residual connections. See \citet{strubell2018linguistically} for more details on these layers. In our experiments we use $K=2$ and convolutional filters of size 1024. We use the typical 100 dimensional GloVe embeddings pre-trained on 6 billion tokens of text from Wikipedia and Gigaword.\footnote{\protect\url{https://nlp.stanford.edu/projects/glove/}}

\subsection{ELMo embedding model \label{sec:elmo}}
The ELMo model \citep{peters2018deep} produces deep, contextually embedded token representations by training stacked convolutional, highway and bi-directional LSTM (bi-LSTM) layers on a large corpus of text with an unsupervised language modeling (LM) objective. The expressiveness of this model compared to GloVe-style embeddings models differs in two key ways: (1) ELMo observes the entire sentence of context to model each token rather than relying on a small, fixed window and (2) ELMo does not rely on a fixed vocabulary of token embeddings, instead building up token representations from characters. 

The ELMo architecture enhances the bidirectional LM architecture from \citet{peters2017semi}. The model first composes character embeddings into word type embeddings using a convolutional layer followed by highway layers. Then these token representations are passed to multiple bi-LSTM layers, all of which are trained end-to-end to jointly optimize forward and backward LM objectives. ELMo additionally learns a small number of task-specific parameters to compose and scale the outputs of each LM, producing a task-specific embedding for each token in context. The intuition behind these task-specific parameters is that different tasks will benefit from different weightings of shallower and deeper LM representations. For example, parsing might favor earlier layers which better capture syntactic patterns, whereas question answering might favor later layers which capture higher level semantics of the sentence. \citet{peters2018deep} experiment with adding ELMo embeddings on the input and output of some architectures, with varying results across different tasks. We incorporate ELMo embeddings into the model by keeping the pre-trained parameters fixed but learning a task-specific combination of the layer outputs which we feed as inputs to our model, as described in \citet{peters2018deep}. We follow their implementation for the SRL architecture of \citep{he2017deep} and use the ELMo embeddings only as input to the model. We refer to \citet{peters2018deep} for more details on this model.

We use the pre-trained TensorFlow ELMo model\footnote{\protect\url{https://github.com/allenai/bilm-tf}}, which consists of one character-level convolutional layer with 2048 filters followed by two highway layers followed by two bi-LSTM layers with 4096 hidden units. All three layers are projected down to 512 dimensional representations over which our task-specific parameters are learned. This model is trained on the 1B Word Benchmark \citep{chelba2014one}, which consists of filtered English newswire, news commentary and European parliament proceedings from the WMT '11 shared task.

\section{Related work}
Our experiments are based on the LISA model of \citet{strubell2018linguistically}, who showed that their method for incorporating syntax into a deep neural network architecture for SRL improves SRL F1 with predicted predicates on CoNLL-2005 and CoNLL-2012 data, including on out-of-domain test data. Other recent works have also found syntax to improve neural SRL models when evaluated on data from the CoNLL-2009 shared task: \citet{roth2016neural} use LSTMs to embed syntactic dependency paths, and \citet{marcheggiani2017encoding} incorporate syntax using graph convolutional neural networks over predicted dependency parse trees. In contrast to this work, \citet{marcheggiani2017encoding} found that their syntax-aware model did not out-perform a syntax-agnostic model on out-of-domain data.

The idea that an SRL model should incorporate syntactic structure is not new, since many semantic formalities are defined with respect to syntax. Many of the first approaches to SRL \citep{pradhan2005semantic,surdeanu2007combination,johansson2008dependency,toutanova2008global, punyakanok2008importance}, spearheaded by the CoNLL-2005 shared task \citep{carreras2005introduction}, achieved success by relying on syntax-heavy linguistic features as input for a linear model, combined with structured inference which could also take syntax into account. \citet{tackstrom2015efficient} showed that most of these constraints could more efficiently be enforced by exact inference in a dynamic program. While most techniques required a predicted parse as input, \citet{sutton2005joint} modeled syntactic parsing and SRL with a joint graphical model, and \citet{lewis2015joint} jointly modeled SRL and CCG semantic parsing. \citet{collobert2011natural} were among the first to use a neural network model for SRL, using a CNN over word embeddings combined with globally-normalized inference. However, their model failed to out-perform non-neural models, both with and without multi-task learning with other NLP tagging tasks such as part-of-speech tagging and chunking. \citet{fitzgerald2015semantic} were among the first to successfully employ neural networks, achieving the state-of-the-art by embedding lexicalized features and providing the embeddings as factors in the model of \citet{tackstrom2015efficient}.

Recently there has been a move away from SRL models which explicitly incorporate syntactic knowledge through features and structured inference towards models which rely on deep neural networks to learn syntactic structure and long-range dependencies from the data. \citet{zhou2015end} were the first to achieve state-of-the-art results using 8 layers of bidirectional LSTM combined with inference in a linear-chain conditional random field \citep{lafferty2001conditional}. \citet{marcheggiani2017simple} and \citet{he2017deep} also achieved state-of-the-art results using deep LSTMs with no syntactic features. While most previous work assumes that gold predicates are given, like this work and \citet{strubell2018linguistically}, \citet{he2017deep} evaluate on predicted predicates, though they train a separate model for predicate detection. Most recently, \citet{tan2018deep} achieved the state-of-the art on the CoNLL-2005 and 2012 shared tasks with gold predicates and no syntax using 10 layers of self-attention, and on CoNLL-2012 with gold predicates \citet{peters2018deep} increase the score of \citet{he2017deep} by more than 3 F1 points by incorporating ELMo embeddings into their model, out-performing ensembles from \citet{tan2018deep} with a single model. We are interested in analyzing this relationship further by experimenting with adding ELMo embeddings to models with and without syntax in order to determine whether ELMo can replace explicit syntax in SRL models, or if they can have a synergistic relationship. 


\section{Experimental results}

In our experiments we assess the impact of replacing GLoVe embeddings ({\bf +GloVe}) with ELMo embeddings ({\bf +ELMo}) in strong, end-to-end neural network models for SRL: one which incorporates syntax ({\bf LISA}) and one which does not ({\bf SA}). The two models are identical except that the latter does not have an attention head trained to predict syntactic heads. Since the LISA model can both predict its own parses as well as consume parses from another model, as in \citet{strubell2018linguistically} we experiment with providing syntactic parses from a high-quality dependency parser ({\bf +D\&M}), as well as providing the gold parses ({\bf +Gold}) as an upper bound on the gains that can be attained by providing more accurate parses. 

We compare LISA models to two baseline models: The deep bi-LSTM model of \citet{he2017deep} and the deep self-attention model of \citet{tan2018deep}. Though both also report ensemble scores, we compare to the single-model scores of both works. We note that \citet{tan2018deep} is not directly comparable because they use gold predicates at test time. Despite this handicap, our best models obtain higher scores than \citet{tan2018deep}.

\subsection{Data and pre-processing}
We evaluate our models on the data from the CoNLL-2005 semantic role labeling shared task \citep{carreras2005introduction}. This corpus annotates the WSJ portion of the Penn TreeBank corpus \citep{marcus1993building} with semantic roles in the PropBank style \citep{palmer2005proposition}, plus a challenging out-of-domain test set derived from the Brown corpus \citep{francis1964manual}. This dataset contains only verbal predicates and 28 distinct role label types. We obtain 105 SRL labels (including continuations) after encoding predicate argument segment boundaries with BIO tags. We use Stanford syntactic dependencies v3.5.

\subsection{Syntactic parsing}
Table~\ref{table:parse} presents the accuracy (UAS) of our dependency parsers. We experiment with adding ELMo embeddings to a strong graph-based dependency parser \citep{dozat2017deep}, and present LISA's parsing accuracy with GloVe and ELMo embeddings. ELMo increases parsing accuracy in both models and both datasets, though by a much wider margin (3 points) for LISA, which attains much lower scores without ELMo. Incorporating ELMo into LISA is very beneficial to parsing accuracy, helping to close the gap between LISA and D\&M parsing performance. In subsequent experiments, D\&M refers to the best model in Table~\ref{table:parse}, D\&M+ELMo.

\begin{table}
\begin{tabular}{lrrrrrr}


& \multicolumn{2}{c}{WSJ Test} && \multicolumn{2}{c}{Brown Test} \\ \cline{2-3} \cline{5-6} 
& GLoVe & ELMo && GLoVe & ELMo \\ \hline
D\&M & 96.13 & 96.48 && 92.01 & 92.56 & \\
LISA & 91.47 & 94.44 && 88.88 & 89.57 & \\
\end{tabular}
\caption{Dependency parse accuracy (UAS) on CoNLL-2005. \label{table:parse}}
\end{table}


\subsection{Semantic role labeling}

\begin{table*}
\begin{tabular}{lrrrrrrrrrrr}
\multirow{2}{*}{Model} & \multicolumn{3}{c}{Dev} && \multicolumn{3}{c}{WSJ Test} && \multicolumn{3}{c}{Brown Test} \\ \cline{2-4} \cline{6-8} \cline{10-12}
& P & R & F && P & R & F && P & R & F \\ \hline \hline
\citet{he2017deep} & 80.3 & 80.4 & 80.3 && 80.2 & 82.3 & 81.2 && 67.6 &  69.6 & 68.5 \\
\citet{tan2018deep}$^\dagger$ & 82.6 & {\bf 83.6} & 83.1 && 84.5 & {\bf 85.2}&  84.8&& 73.5 & {\bf 74.6} & 74.1 \\ \hline
SA+GloVe &  78.54  &  76.90  & 77.71 && 81.43 &  80.69 &  81.06 &&  70.10  & 66.01  & 67.99 \\
LISA+GloVe & 81.25 &  80.03 &  80.64  && 82.78 &  82.57  & 82.68  && 71.93  & 69.45 &  70.67 \\
\ \ \ \ +D\&M & 82.68 &  82.12  & 82.40  && 84.12 &  83.92 &  84.02 && 73.96 &  70.97 &  72.43 \\ 
\ \ \ \ \emph{+Gold} & \emph{86.02} &  \emph{85.11} &  \emph{85.56}  && ---  & --- &  ---  && ---  & --- &  ---\\ \hline
SA+ELMo & 83.67 &  82.37 &  83.02  && 84.29 &  83.95 &  84.12  && 73.76 &  71.02  & 72.36 \\
LISA+ELMo & 84.18 &  82.71  & 83.44  && 84.62 &  84.24 &  84.43  && 73.70 &  71.70 &  72.69 \\
\ \ \ \ +D\&M & {\bf 84.56} &  83.29 &  {\bf 83.92}  && {\bf 85.40} &  84.93 &  {\bf 85.17}  && {\bf 75.27}  & 73.40 &  {\bf 74.33} \\
\ \ \ \ \emph{+Gold} & \emph{87.56}  & \emph{86.01}  & \emph{86.77}  && --- & ---  & ---  && --- & --- & --- \\
\end{tabular}
\caption{Precision, recall and F1 on CoNLL-2005 with predicted predicates. $\dagger$ denotes that models were evaluated on gold predicates. \label{table:srl}}
\end{table*}

In Table~\ref{table:srl} we present our main results on SRL. First, we see that adding ELMo embeddings increases SRL F1 across the board. The greatest gains from adding ELMo are to the SA models, which do not incorporate syntax. ELMo embeddings improve the SA models so much that they nearly close the gap between SA and LISA: with GloVe embeddings LISA obtains 1.5-2.5 more F1 points than SA, whereas with ELMo embeddings the difference is closer to 0.3 F1. This is despite ELMo embeddings increasing LISA's parse accuracy by 3 points on the WSJ test set (Table~\ref{table:parse}). These results suggest that ELMo does model syntax in some way, or at least the model is able to leverage ELMo embeddings about as well as LISA's predicted parses to inform its SRL decisions. 

\includegraphics[scale=0.0000001]{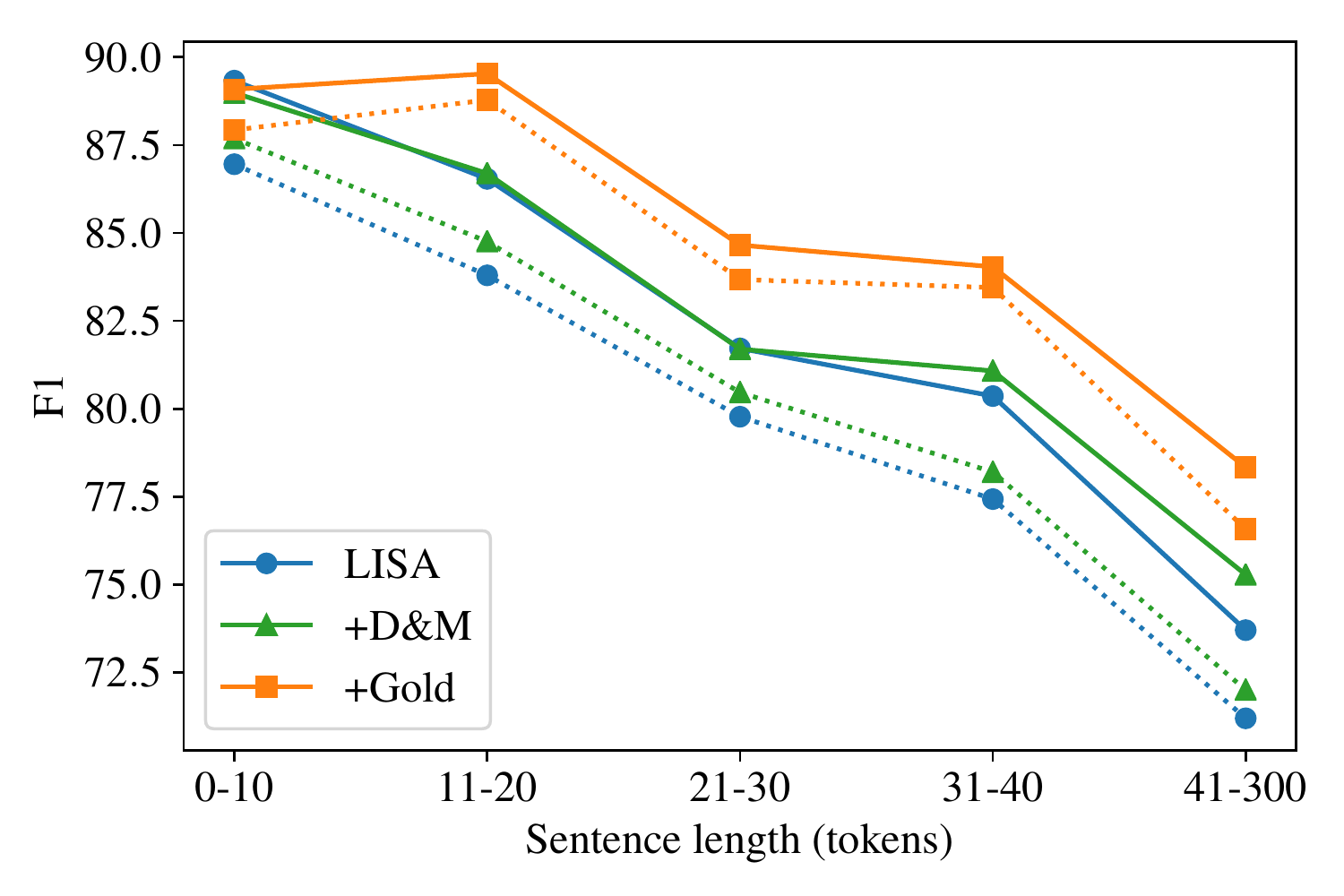}

However, when we add higher-accuracy parses (+D\&M) to LISA, we do see greater improvements over the syntax-agnostic model, even with ELMo embeddings. We see that incorporating explicit syntax representations is still helpful even with ELMo's strong representations. On the WSJ test set, supplying LISA with D\&M parses gives about 1 point of F1 over the SA baseline, and on the out-of-domain test set, we see that the parses supply almost 2 additional points of F1 over the syntax-agnostic model. We note that with ELMo embeddings and D\&M parses, LISA obtains new state-of-the-art results for a single model on this dataset \emph{when compared to a model \citep{tan2018deep} which is given gold predicates at test time}, despite our models using predicted predicates. Our model's gains in F1 come from obtaining higher precision than \citet{tan2018deep} (fewer false positives). 

The explicit parse representations appear to be particularly helpful on out-of-domain data, which makes sense for two reasons: First, since the Brown test set is out-of-domain for the ELMo embeddings, we would expect them to help less on this corpus than on the in-domain WSJ text. Second, the parse trees should provide a relatively domain-agnostic signal to the model, so we would expect them to help the most in out-of-domain evaluation.

We also evaluate on the development set with gold parse trees at test time. Fairly large gains of nearly 3 F1 points can still be obtained using gold parses even with ELMo embeddings, suggesting that syntax could help even more if we could produce more accurate syntactic parses, or more specifically, the types of mistakes still made by highly accurate dependency parsers (e.g. prepositional phrase attachments) negatively impact SRL models which rely on syntax. 

\begin{table}
\begin{tabular}{llrrr}
& & \multicolumn{3}{c}{WSJ Test} \\ \cline{3-5}
& & P & R & F \\ \hline \hline
\multicolumn{2}{c}{\citet{he2017deep}} & 94.5 & 98.5 & 96.4 \\ \hline
\multirow{2}{*}{GloVe} & SA &  98.27 &  98.14 &  98.20 \\
& LISA & 98.34 &  98.04 &  98.19  \\ \hline
\multirow{2}{*}{ELMo} & SA & 98.66 &  97.51 &  98.08 \\
& LISA & 98.58 &  97.28 &  97.93 \\
& & & & \\
& & \multicolumn{3}{c}{Brown Test} \\ \cline{3-5}
& & P & R & F \\ \hline \hline
\multicolumn{2}{c}{\citet{he2017deep}} & 89.3 & 95.7 & 92.4\\ \hline
\multirow{2}{*}{GloVe} & SA &  94.68  & 92.91 &  93.79 \\
& LISA & 95.43 &  93.41 &  94.41 \\ \hline
\multirow{2}{*}{ELMo} & SA & 95.70  & 91.42 &  93.51 \\
& LISA & 96.46 &  91.54 &  93.94 \\
\end{tabular}
\caption{Predicate detection precision, recall and F1 on CoNLL-2005. \label{table:predicates}}
\end{table}

Table~\ref{table:predicates} lists precision, recall and F1 of our predicate detection. We note that there is very little difference in predicate detection F1 between GloVe and ELMo models, demonstrating that the difference in scores can not be attributed to better predicate detection. If anything, the predicate detection scores with ELMo are slightly lower than with GloVe. We observe that in particular on the Brown test set, ELMo predicate detection precision is notably higher than GloVe while recall is lower.

\subsection{Analysis}
\begin{figure}
\includegraphics[scale=0.5]{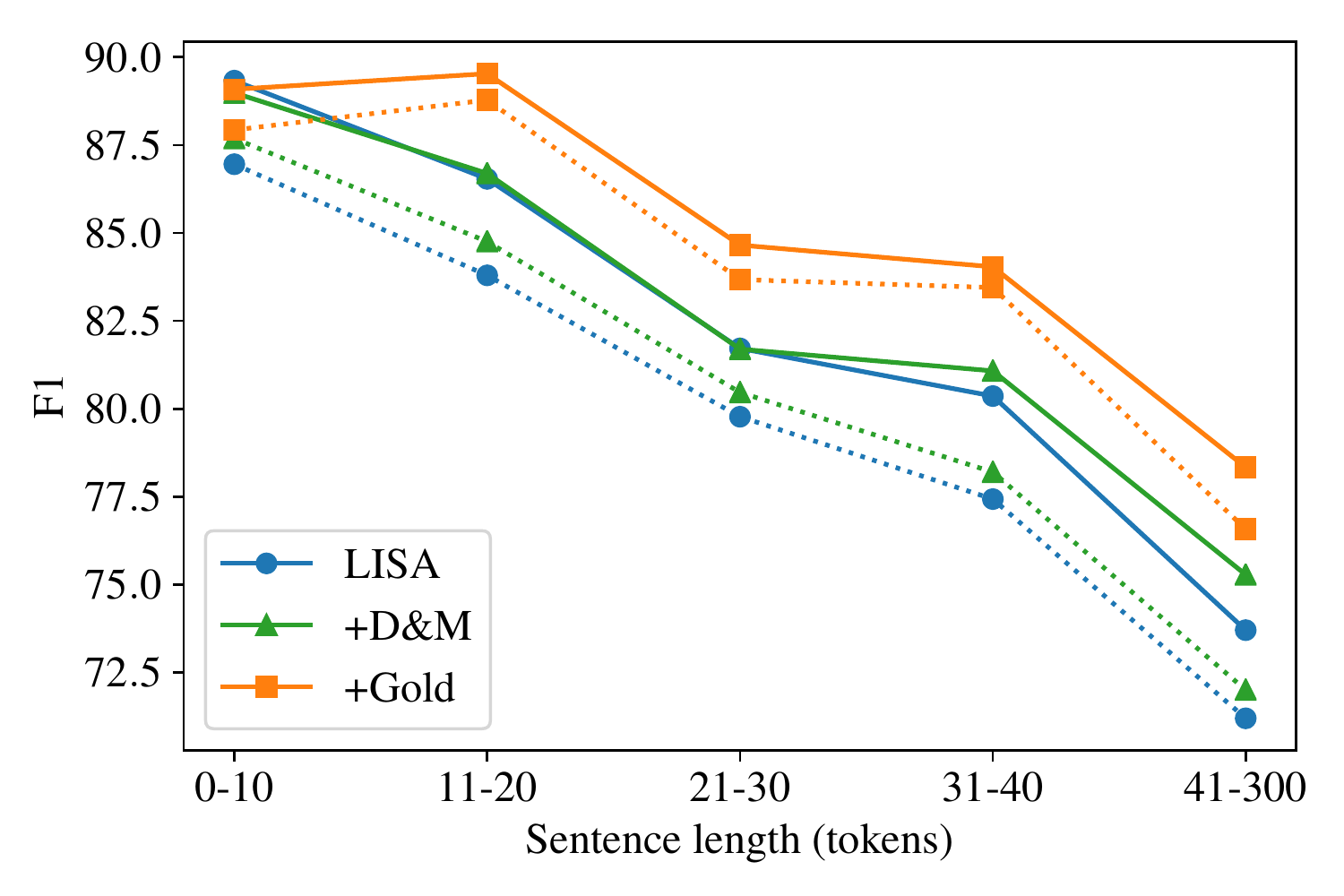}
\caption{F1 score as a function of sentence length. Solid/dotted lines indicate ELMo/GLoVe embeddings, respectively.\label{fig:sent-lens}}
\end{figure}

We follow \citet{strubell2018linguistically} and \citet{he2017deep} and perform an analysis on the development dataset to ascertain which types of errors ELMo helps resolve, and how this compares with the types of errors that occur when LISA is provided with a gold parse. In Figure~\ref{fig:sent-lens} we bucket sentences by length and plot F1 score as a function of sentence length across different models reported in Table~\ref{table:srl}. Solid lines indicate ELMo models, while dotted lines indicate models trained with GloVe. For GloVe models, we see that models without gold parses maintain about the same difference in F1 across all sentence lengths, while the gold parse obtains significant advantages for sentence lengths greater than 10. With ELMo embeddings, the relationship between models with gold and predicted parses remains the same, but interestingly the LISA and D\&M parses obtain the same scores through sentence lengths 21-30, then diverge more as sentences get longer. This trend suggests that while the ELMo embeddings help LISA parse shorter sentences, up to length 30, the D\&M model trained specifically on parsing is more accurate on longer sentences. This could be indicative of ELMo's ability to model long-range dependencies. 

\begin{figure}
\includegraphics[scale=0.5]{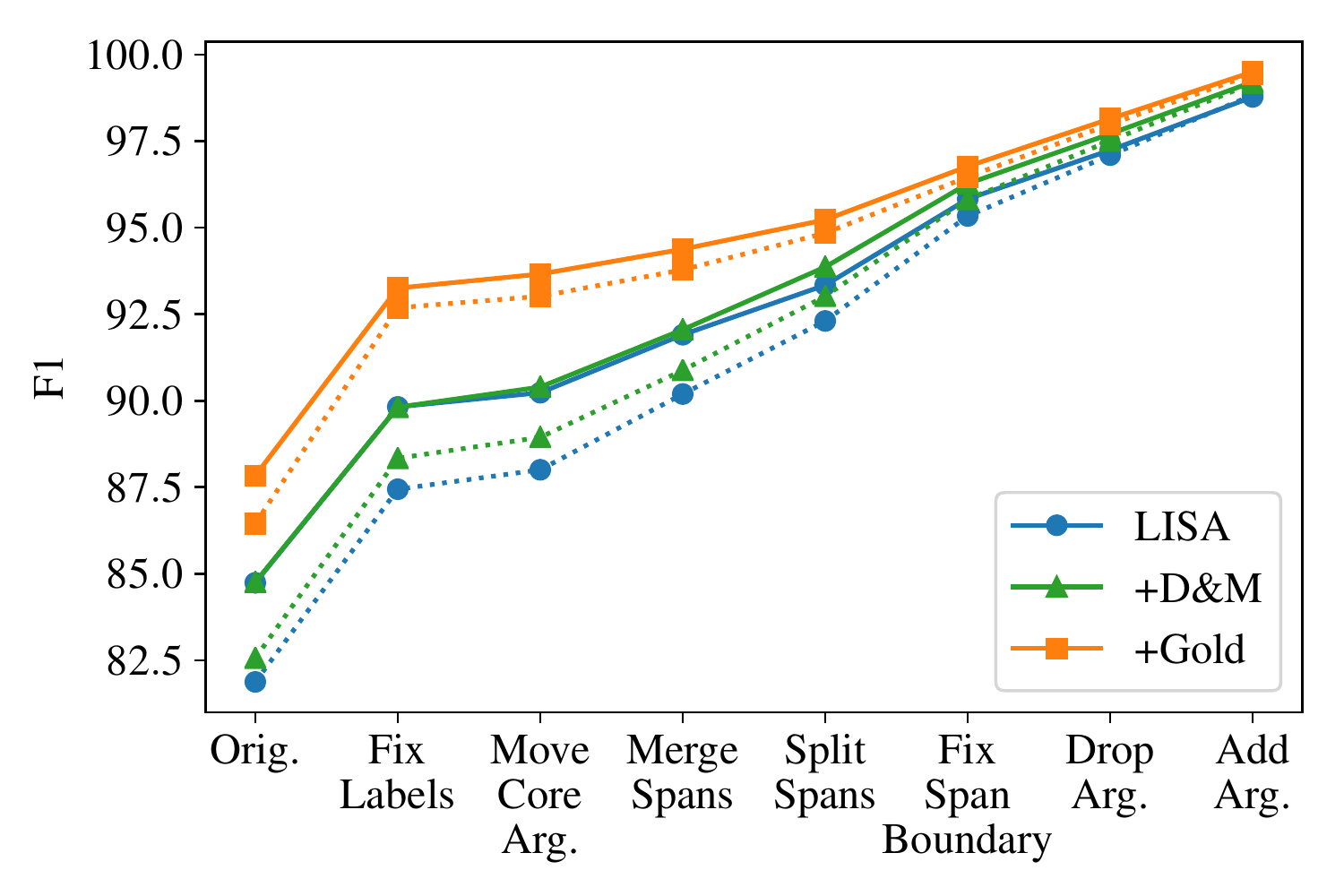}
\caption{F1 score on CoNLL-2005 after performing incremental corrections from \citet{he2017deep}. Solid/dotted lines indicate ELMo/GLoVe embeddings, respectively. \label{fig:error-types}}
\end{figure}

In Figure~\ref{fig:error-types} we follow the analysis from \citet{he2017deep} which bins SRL errors into 7 different error types,\footnote{Refer to \citet{he2017deep} for more detailed descriptions of the error types} then incrementally fixes each error type in order to better understand which error types contribute most to SRL errors. We follow \citet{strubell2018linguistically} and compare these errors across models with access to different levels of parse information, and which are trained with GloVe and ELMo word representations. As in Figure~\ref{fig:sent-lens}, solid lines in Figure~\ref{fig:error-types} represent models trained with ELMo embeddings and the dashed lines indicate models trained with GloVe.

The overall trend is that supplying the gold parse helps most with segment boundary mistakes, i.e. those resolved by merging or splitting predicted role spans, for both ELMo and GloVe models. The ELMo models clearly begin to close the gap between models given predicted and gold parses by making less of these boundary mistakes, which is not simply due to better parse accuracy since the GlovE+D\&M model has access to the same parses as ELMo+D\&M.

\section{Conclusion}
To address the question of whether syntax is still relevant in SRL models with tokens embedded by deep, unsupervised, sentence-aware models such as ELMo, we compared the performance of LISA, a syntactically-informed SRL model, trained with ELMo and GloVe token representations. We found that although these representations improve LISA's parsing and SRL tagging performance substantially, models trained to leverage syntax still obtain better F1 than models without syntax even when provided with ELMo embeddings, especially on out-of-domain data. We conclude that syntax is indeed still relevant in neural architectures for SRL. In future work, we are interested in exploring similar analysis for NLP tasks which have less obvious ties to syntactic structure.

\section*{Acknowledgments}
We thank Luheng He for providing her excellent error analysis scripts, Timothy Dozat and the authors of tensor2tensor for releasing their code, Daniel Andor and David Weiss for helpful discussions, and the reviewers for their thoughtful comments. This work is supported in part by the Center for Data Science and the Center for Intelligent Information
Retrieval, in part by the Chan Zuckerberg
Initiative under the project ``Scientific Knowledge
Base Construction,'' and in part by an IBM PhD
Fellowship Award to ES. Any opinions, findings and conclusions
or recommendations expressed in this material
are those of the authors and do not necessarily
reflect those of the sponsor.

\bibliography{acl-ws}
\bibliographystyle{acl_natbib}

\end{document}